\documentclass[a4paper, 11pt]{article}
\usepackage[utf8]{inputenc}
\usepackage[brazil]{babel}
\usepackage{tabularx}
\usepackage{cite}
\usepackage{subfigure}
\usepackage{hyperref}
\usepackage{multirow}
\usepackage[table]{xcolor}
\usepackage{color}
\usepackage{graphicx}

\usepackage{amssymb}

\usepackage{xcolor,colortbl}

\newcommand{\best}[1]{\cellcolor{gray}\textcolor{white}{\textbf{{#1}}}}

\usepackage{xcolor,colortbl}

\title{Um arcabouço de Seleção de Bandas Landsat-8 baseado em UMDA para Detecção de Desmatamento}
\author{Eduardo B. Neto, Paulo R. C. Pedro, Álvaro Fazenda,  Fabio A. Faria\\
Universidade Federal de São Paulo -- São José dos Campos, Brasil, 
\\ \{ebneto, costa.paulo, alvaro.fazenda, ffaria\}@unifesp.br}

\begin{document}

\maketitle
\begin{abstract}
 A conservação das florestas tropicais é um assunto atual de relevância social e ecológica, devido ao importante papel que elas desempenham no ecossistema global. Infelizmente, milhões de hectares são desmatados e degradados todo ano, sendo necessários programas (governamentais ou de iniciativas privadas) para monitoramento das florestas tropicais. Neste sentido, este trabalho propõe um novo arcabouço que utiliza de algoritmo de estimação de distribuição estocástico (UMDA) para selecionar bandas espectrais do satélite Landsat-8 que melhor representam as áreas de desmatamento que guiará a criação de novas composições de imagens para alimentar uma arquitetura de segmentação semântica chamada DeepLabv3$+$. Nos experimentos realizados foi possível encontrar diversas composições que conseguem acurácia balanceada maior que 90\% na tarefa de classificação de segmentos. Além disso, a melhor composição de bandas ($651$) encontrada  pelo algoritmo UMDA alimentou uma arquitetura DeepLabv3$+$ e resultou em melhor eficiência e eficácia contra todas as composições comparadas neste trabalho\footnote{\textbf{Códigos Fonte disponível em:} \url{https://github.com/ebouhid/}.}. 
\end{abstract}
\clearpage

\section{Introdução}

Florestas tropicais estão localizadas entre os Trópicos de Câncer e Capricórnio, próximas à linha do Equador, sendo encontradas na América do Sul e Central, na África e em regiões da Ásia e do Pacífico. Até $2015$ existiam $700$ milhões de hectares de floresta primária (original)~\cite{martin2015edge}.   Embora as florestas tropicais cubram apenas $7$\% da superfície terrestre, estima-se que elas abrigam mais da metade das espécies do planeta~\cite{martin2015edge}. Além da grande biodiversidade, as plantas e o solo das florestas tropicais retêm de $460$ a $575$ bilhões de toneladas de carbono. Os processos de evaporação e evapotranspiração de suas plantas e árvores retornam grandes quantidades de água para a atmosfera local, promovendo a formação de nuvens e chuvas~\cite{nasaflorestas}.

Infelizmente, milhões de hectares de florestas tropicais tem sido perdidos a cada ano através de desmatamento e degradação~\cite{Hansen850,martin2015edge}. De acordo com um dos mais conhecidos e bem sucedidos programas de monitoramento, PRODES (Programa de
Monitoramento da Floresta Amazônica Brasileira por Satélite), no período entre agosto/$2021$ e julho/$2022$, o desmatamento na Amazônia Legal Brasileira alcançou $11.568$ $km^2$ \cite{prodes2022}. E para piorar o cenário, com a escassez de mão-de-obra especializada e abundância de dados a serem analisados, o processo de monitoramento da floresta se torna muito custoso tanto financeiramente quanto em relação ao tempo, o que constitui um grande desafio para as tecnologias da informação e comunicação (TIC).

Na literatura, técnicas de aprendizado de máquina e processamento digital de imagens têm sido desenvolvidas para superar os desafios encontrado e auxiliar os especialistas nas tarefas de detecção automática ou semi-automática de áreas de desmatamento em florestas tropicais. Ortega \textit{et al.}~\cite{ortega2019evaluation}, aponta que o uso de métodos de classificação baseados em aprendizado profundo – particularmente, Redes Neurais Convolucionais Siamesas e \textit{Early Fusion} – é uma abordagem superior a técnicas tradicionais de Aprendizado de Máquina. Andrade \textit{et al.}~\cite{andrade2020evaluation} aborda a detecção de desmatamento em imagens de satélite a partir da Segmentação Semântica (SS) utilizando o modelo DeepLabv3$+$. Ambos trabalhos utilizam imagens do satélite Landsat-8 com todas as sete bandas espectrais com adição de NDVI. Em Maretto \textit{et al.}~\cite{maretto2020spatio}, foi proposta uma abordagem para detecção de desmatamento via segmentação semântica de imagens do Landsat-8 com fusão de imagens baseada na conhecida arquitetura U-Net. Para isso, foram utilizadas imagens compostas por 5 bandas (3, 4, 5, 6 e 7), escolhidas ad-hoc.

No sentido de explorar o real potencial de sensores multiespectrais tais como o Landsat-8, muitos trabalhos têm sido propostos para criar novas imagens de uma mesma área por meio da escolha de bandas espectrais e buscar as melhores representações para aplicação alvo. Yu \textit{et al.}~\cite{yu2019selection} propõe uma metodologia de seleção de bandas baseada em suas correlações para otimizar o desempenho de uma Máquina de Vetores de Suporte (SVM) na tarefa de classificação do uso/cobertura de terras. Já Dallaqua et al.~\cite{foresteyes2019}, realizou um estudo ``força bruta"$\ $ na busca da melhor composição de bandas que proporcionasse maior acurácia na detecção de desmatamento por um classificador SVM. Nesse contexto, o uso de algoritmos de busca se mostra como uma boa solução alternativa, por ser mais eficiente em tempo e custo computacional do que métodos tradicionais, como \textit{grid search} convencional~\cite{ma2003bandselection}. Dentre os trabalhos existentes na literatura que abordam esse tema destaca-se Nagasubramanian~\textit{et al.}~\cite{nagasubramanian2018hyperspectral_ga} que faz uso de algoritmos genéticos na busca da melhor combinação de bandas de um sensor hiperespectral para a detecção de pragas em plantações de soja.

Nesse sentido, este trabalho propõe um arcabouço de seleção de bandas espectrais baseado em UMDA (\textit{Univariate Marginal Distribution Algorithm}) para servir como guia para a construção da base de imagens de entrada para arquiteturas de segmentação semântica de imagens de desmatamento da Amazônia.

\section{Um arcabouço de Seleção de Bandas LandSat-8 baseado em UMDA para Detecção de Desmatamento}
\label{sec:fundamentos}
\begin{figure*}[ht!]
    \centering
    \includegraphics[width=1\linewidth]{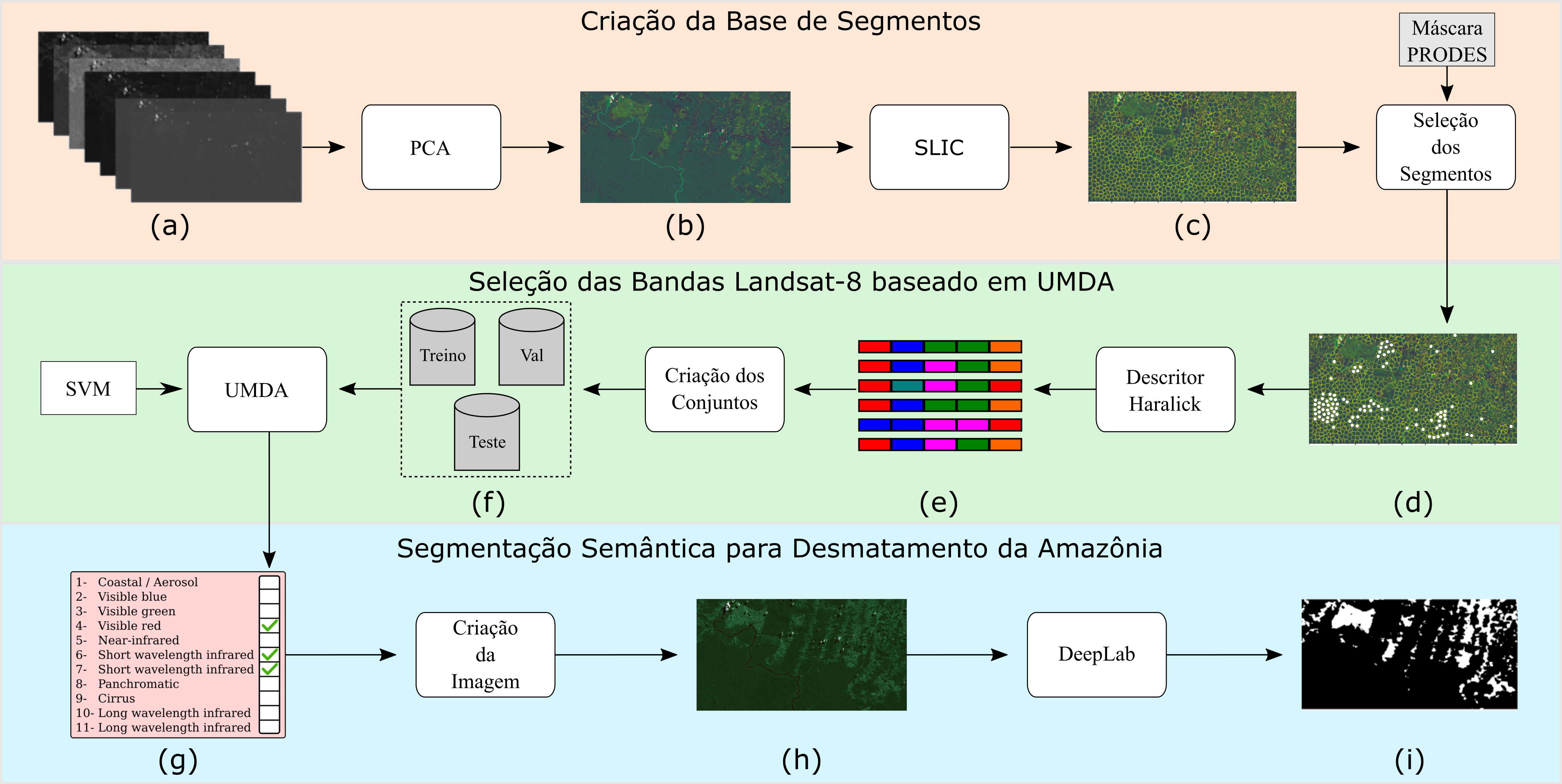}
    \caption{O arcabouço de seleção de bandas Landsat-8 baseado em UMDA para detecção de desmatamento.}
    \label{fig:workflow}
\end{figure*}

A Figura~\ref{fig:workflow} mostra em detalhes o funcionamento do arcabouço proposto neste trabalho. Em (a) está uma imagem do satélite Landsat-8 com suas sete bandas espectrais, (b) é a imagem falsa-cor formada de três componentes principais do PCA~\cite{PCA},  (c) é a imagem resultante da segmentação SLIC~\cite{SLIC} com diferentes segmentos encontrados, (d) é a imagem resultante da seleção de segmentos indicados pelos pontos brancos, (e) estão os vetores de características Haralick~\cite{HARALICK} extraídos dos segmentos selecionados, (f) estão os vetores de características separados em três conjuntos (treino, validação e teste) para serem utilizados no treinamento do classificador SVM~\cite{svm} durante processo evolutivo do algoritmo UMDA~\cite{muhlenbein1996recombination}, (g) está o melhor indivíduo encontrado pelo algoritmo UMDA (bandas 4, 6 e 7), (h) está a composição da imagem representada pelo melhor indivíduo e finalmente em (i) está a imagem segmentada pela arquitetura DeepLabv3$+$~\cite{chen2018encoderdecoder}, onde os \textit{pixels} brancos correspondem às regiões de desmatamento na imagem de entrada. 

\subsection{Criação da Base de Segmentos}
\label{sec:dataset}
    A base de dados criada para este trabalho é composta por 10 diferentes imagens do satélite Landsat-8 da região da Bacia do Xingu que totaliza mais de 9.497 $km^2$ de floresta. Para a realização dos experimentos foi adotado o mesmo protocolo experimental de~\cite{foresteyes2019}. Portanto,   
    cada uma das $10$ imagens composta por $7$ bandas espectrais foram submetidas ao algoritmo de redução de dimensionalidade PCA passando de 7 para apenas 3 bandas (3 primeiras componentes principais) e um algoritmo de segmentação não-supervisionada SLIC~\cite{SLIC} foi aplicado para criação de 11831 segmentos (11404 de ``Floresta" e 427 de ``Não-floresta"). Uma vez com segmentos criados, realizou-se uma etapa de seleção dos melhores segmentos com adoção de dois critérios: (1) segmentos com Taxa de Homogeneidade (proporção da classe majoritária) $ \geq 0,70$ considerando a verdade fornecida pelo PRODES~\cite{prodes2022}; e (2) segmentos com área mínima de 70 \textit{pixels}. 
    A Tabela ~\ref{tab:datasettotal} descreve as $10$ regiões com diferentes dimensões e suas representações em km$^2$. Os valores de Floresta e Não-floresta estão relacionados com a quantidade de segmentos encontrados pelo SLIC para cada uma das classes.   
\vspace{-0.25cm}
    \begin{table}[ht!]
    \centering
    \caption{Descrição detalhada das imagens Landsat-8 coletadas para  este trabalho. }
    \begin{tabular}{|c|c|c|c|c|c|}
    \hline
        \textbf{Região} & \textbf{km$^2$} &\textbf{Dimensões} &\textbf{Floresta} & \textbf{Não-floresta} & \textbf{Total} \\ \hline
        {1} & 930,40 &1230$\times$ 843 & 1585  & 107  & 1692  \\ \hline
        {2} & 1124,59 & 1343$\times$ 933 & 1115  & 134  & 1249  \\ \hline
        {3} & 875,36 & 977$\times$ 998 & 1167  & 25  & 1192  \\ \hline
        {4} & 607,33 & 768$\times$ 879 & 475  & 9  & 484  \\ \hline
        {5} & 982,86 & 790$\times$ 1384 & 612  & 36  & 648  \\ \hline
        {6} & 694,77 & 928$\times$ 833 & 1104  & 17  & 1121  \\ \hline
        {7} & 1528,65 & 1788$\times$ 950 & 2016  & 9  & 2025  \\ \hline
        {8} & 778,84 & 853$\times$ 1017 & 928  & 36  & 964  \\ \hline
        {9} & 925,91 & 971$\times$ 1064 & 1582  & 19  & 1601  \\ \hline
        { 10} & 1048,45 & 1047$\times$ 1115 & 820  & 35  & 855  \\ \hline \hline
         \textbf{Total} &$\sim$\textbf{9.497}& -- & \textbf{11404} &  \textbf{427} &  \textbf{11831}\\\hline
    \end{tabular}
    \label{tab:datasettotal}
\end{table}

\subsection{Seleção das Bandas Landsat-8 baseada em UMDA}

Para cada um dos $11831$ segmentos, são calculados seus respectivos descritores de textura Haralick~\cite{HARALICK} compostos de $13$ coeficientes calculados em $4$ direções para cada uma das $7$ bandas espectrais, o que totaliza $7$ vetores de características para cada segmento. Então, os segmentos são separados em três conjuntos ($70$\% de treino, $15$\% de validação e $15$\% de teste), os quais serão utilizados no processo evolutivo do algoritmo UMDA.  
Introduzido por Mühlenbein \& Paaß~\cite{muhlenbein1996recombination}, o algoritmo \textit{Univariate Marginal Distribution} (UMDA) é um dos mais simples algoritmos de estimação de distribuição (AED) que assumem independência entre variáveis de decisão e tem obtido resultados superiores aos alcançados pelos algoritmos genéticos tradicionais~\cite{Ferreira_icpr2020}. 
Em seu processo evolutivo, cada indivíduo da população é composto de 7 genes que representam as bandas espectrais que compõem uma imagem Landsat-8. A presença da banda no indivíduo é marcada com valor $1$ e sua ausência com $0$. O algoritmo inicia com uma população aleatória, seguido de cálculo da função de aptidão de cada indivíduo. A função de aptidão utilizada é a acurácia balanceada de um classificador SVM (\textit{Support Vector Machine}), empregando segmentos formados pelos descritores de Haralick referentes às bandas do indivíduo em questão. Em seguida, os $n$ melhores indivíduos são selecionados para compor o conjunto de indivíduos ``pais". As probabilidades marginais são calculadas a partir da frequência relativa de cada gene presente entre todos os pais, estimando sua distribuição de probabilidade. Considerando essa distribuição de cada gene, novos indivíduos são gerados como ``prole". A população da próxima geração é composta pela união dos conjuntos de indivíduos da ``prole" e dos ``pais". O processo é repetido até que o critério de parada seja atingido, sendo a quantidade de gerações o critério adotado neste estudo. Dentre as diversas configurações testadas para o algoritmo UMDA  visando eficiência no processo evolutivo, os melhores parâmetros encontrados foram:  $10$ indivíduos na população, $10$ gerações, número de indivíduos selecionados (pais) igual a $5$ e número descendentes gerados igual a $5$.

\subsection{Segmentação Semântica para Desmatamento da Amazônia}
 Nesta etapa, as $10$ diferentes imagens originais presentes na base de dados Landsat-8, descritas na Tabela~\ref{tab:datasettotal}, são reconstruídas por meio da composição das bandas mais relevantes (isto é, aquelas com maior representatividade dentre os melhores indivíduos do UMDA, conforme a Tabela~\ref{tab:best_individuals}). Por sua vez, essas imagens são separadas em dois conjuntos, sendo que o treino é composto de 8 regiões (1, 2, 5, 6, 7, 8, 9 e 10) e o teste composto por 2 regiões (3 e 4). Similar ao trabalho~\cite{andrade2020evaluation}, essa configuração dos conjuntos de treino e teste se deram de forma manual, uma vez que é desejável que imagens com nuvens não estivessem presentes no conjunto de teste. 

Durante processo de treinamento da arquitetura de segmentação semântica DeepLabv3$+$~\cite{chen2018encoderdecoder}, cada imagem dos conjuntos de treino e teste é recortada em sub-imagens de dimensões $256 \times 256$ com \textit{stride} de $64$ \textit{pixels}. O recurso de \textit{data augmentation} foi utilizado para o conjunto de treino, que sofreu transformações de rotação, apenas. Com isso, obteve-se $10728$ imagens para treino e $234$ para teste.

\section{Resultados \& Discussão}
\label{sec:results}
Esta seção discute em detalhes os resultados dos experimentos realizados neste trabalho.

\subsection{Classificação de Segmentos pelo Classificador SVM}

Neste experimento, realizou-se uma análise comparativa dos resultados de classificação do SVM utilizando os melhores indivíduos encontrados pelo algoritmo UMDA.

A Tabela~\ref{tab:best_individuals} mostra apenas os indivíduos encontrados pelo algoritmo UMDA com acurácia balanceada maior que $90,0\%$ obtida pelo classificador SVM no conjunto de teste. É possível observar, como em todo algoritmo não-determinístico, os resultados de acurácia dos indivíduos/soluções variam dependendo da semente escolhida no experimento. Entretanto, um fato muito interessante de comentar é a presença frequente das bandas $6$, $5$ e $1$, observadas $92,3\%$, $92,3\%$ e $69,2\%$ dos indivíduos, respectivamente. Portanto, pode-se sugerir a grande importância dessas bandas para a tarefa alvo.

A Tabela~\ref{tab:best_individuals_baselines} mostra uma análise comparativa entre os melhores indivíduos encontrados pelo algoritmo UMDA e três composições de bandas comumente utilizadas na literatura com as cinco diferentes sementes de inicialização. É possível notar que os resultados do classificador SVM utilizando os melhores indivíduos conseguem resultados similares aos da melhor composição da literatura, a qual utiliza de todas as sete bandas espectrais. Já quando comparados os melhores indivíduos com as composições RGB e PCA, os ganhos relativos são $11,88\%$ e $5,93\%$, respectivamente.

\subsection{Segmentação Semântica de Imagens de Desmatamento da Amazônia}
Visando enfatizar a importância das bandas espectrais mais frequentes encontradas no experimento anterior para a tarefa de classificação de segmentos de desmatamento na Amazônia, a Tabela~\ref{tab:semantica} apresenta uma análise comparativa da segmentação semântica usando a arquitetura DeepLabv3$+$ com diferentes variações das bandas mais frequentes. Em todos os experimentos, utilizamos como base a arquitetura DeepLabv3$+$ com os pesos do \textit{encoder} pré-treinados na base de dados ImageNet~\cite{deng2009imagenet}. Em seguida, realizamos o ajuste fino do modelo com os seguintes parâmetros: 100 épocas de treinamento, taxa de aprendizado de $5 \times 10^{-4}$ e função de custo "Jaccard Loss". Além disso, outras três composições da literatura foram adicionadas à análise (todas$+$NDVI~\cite{torres_2021}, PCA e RGB).
Como pode ser observado, o melhor resultado na tarefa alvo foi obtido pela composição das três bandas espectrais mais frequentes do estudo anterior (6, 5 e 1), conseguindo superar todas as três composições da literatura comparadas neste experimento, inclusive a composição de todas as 7 bandas do Landsat-8 + NDVI proposta por~\cite{torres_2021} para 3 de 4 medidas de avaliação calculadas e, também, no tempo de convergência da arquitetura DeepLabv3$+$. Este fato sugere que a seleção de bandas espectrais baseada em UMDA pode trazer grandes benefícios de eficiência (ganho de tempo) e eficácia (qualidade de resultados) para a tarefa de segmentação semântica de imagens de desmatamento da Amazônia.

\begin{table}
    \centering
    \caption{Resultados do classificador SVM com os melhores indivíduos encontrados pelo algoritmo UMDA. O símbolo \checkmark significa a presença da banda espectral no indivíduo.}
    \resizebox{13cm}{!}{
    \begin{tabular}{|c|c|c|c|c|c|c|c|c|c|} \hline
\multirow{2}{*}{\textbf{Semente}} & \textbf{Indivíduo} & \multicolumn{7}{c|}{\textbf{Bandas}} & \textbf{Acurácia}\\  \cline{3-9}
 & \textbf{UMDA} & \textbf{1} & \textbf{2} & \textbf{3} & \textbf{4} & \textbf{5} & \textbf{6} & \textbf{7} & \textbf{Balanceada}\\  \hline
10 & 1 & \checkmark & & \checkmark & \checkmark & \checkmark & \checkmark & & 93,74\\ \hline
 42 & 2 & & \checkmark & & \checkmark & \checkmark & \checkmark & & 93,24\\ \hline
 30 & 3 & \checkmark & & \checkmark & \checkmark & \checkmark & \checkmark & \checkmark & 93,12\\ \hline
 30 & 4 & \checkmark & & \checkmark & & \checkmark & \checkmark & \checkmark & 92,71\\ \hline
 20 & 5 & & \checkmark & \checkmark & \checkmark & \checkmark & \checkmark & \checkmark & 92,67\\ \hline
 20 & 6 & \checkmark & & \checkmark & \checkmark & \checkmark & \checkmark & \checkmark & 92,58\\ \hline
 42 & 7 & & & \checkmark & \checkmark & \checkmark & \checkmark & & 92,51\\ \hline
 10 & 8 & \checkmark & & & & \checkmark & \checkmark & \checkmark & 92,35\\ \hline
 42 & 9 & \checkmark & & & & \checkmark & & \checkmark & 91,63\\ \hline
 1 & 10 & \checkmark & & \checkmark & \checkmark & \checkmark & \checkmark & & 91,50\\ \hline
 1 & 11 & & & & \checkmark & \checkmark & \checkmark & \checkmark & 91,39\\ \hline
 30 & 12 & \checkmark & & \checkmark & & & \checkmark & \checkmark & 90,69\\ \hline
 20 & 13 & \checkmark & \checkmark & & & \checkmark & \checkmark & & 90,22\\ \hline
 
\textbf{Prob.} & - & \best{69,2\%}	& 23,1\%	&61,5\%	& 61,5\%	& \best{92,3\%} & \best{92,3\%}	& 61,5\%& -\\  \hline
\textbf{Ordem} & - & \textbf{3$^{\circ}$}& \textbf{7$^{\circ}$} & \textbf{4$^{\circ}$} & \textbf{5$^{\circ}$} & \textbf{2$^{\circ}$} & \textbf{1$^{\circ}$} & \textbf{6$^{\circ}$} & -\\  \hline
\end{tabular}
}
\label{tab:best_individuals}
\end{table}

\begin{table}[ht!]
    \centering
    \caption{Resultado dos melhores indivíduos para cada semente e três composições conhecidas da literatura.}
    \resizebox{10cm}{!}{
    \begin{tabular}{|c|c|c|c|c|} \hline
    \multirow{2}{*}{\textbf{Semente}} & \textbf{Indivíduo} & \textbf{Todas} & \multirow{2}{*}{\textbf{RGB}} & \multirow{2}{*}{\textbf{PCA~\cite{foresteyes2019}}}  \\ 
    & \textbf{UMDA}& \textbf{(7 Bandas)}&&\\ \hline
1 & 91,50 & 89,42 & 82,21 & 85,41\\ \hline
10 & 93,74  & 91,76 & 84,44 & 88,47\\ \hline
20 & 92,67  & 93,63 & 80,23 & 86,30\\ \hline
30 & 93,12 & 94,17 & 84,90 & 88,94\\ \hline
42 & 93,24  & 94,83 & 83,82 & 89,84\\ \hline
\textbf{Média} & \best{92,85} & 92,76 & 83,12 & 87,79\\ \hline
\textbf{Desvio Padrão} & \best{0,85}  & 2,19 & 1,91 & 1,86\\ \hline
\end{tabular}
}
\label{tab:best_individuals_baselines}
\end{table}

\begin{table}[!ht]
    \centering
    \caption{Resultados da segmentação semântica utilizando DeepLab.}
    
     \resizebox{13cm}{!}{
    \begin{tabular}{|c|c|c|c|c|c|c|c|}
    \hline
  \textbf{Composição} & \textbf{Precision} & \textbf{Recall} & \textbf{F1-score} & \textbf{Acurácia}&\textbf{mIoU}&\textbf{Época}\\ \hline
      6  & 88,00 & 89,68 & 88,83 & 87,68 & 79,91 & 23 \\ \hline
      6 e 5  & 89,21  & 89,74 & 89,47 & 88,46 & 80,95 & 5 \\\hline
     \best{6, 5 e 1}  & {86,76} & \best{93,22} & \best{89,87} & {88,52} & \best{81,61} & \best{5} \\ \hline
      6, 5, 1 e 3 & 90,41 & 88,30 & 89,34 & 88,49 & 80,74 & 13 \\ \hline
      6, 5, 1 e 4 & 86,83 & 91,29 & 89,00 & 87,68 & 80,19 & 2\\ \hline
      6, 5, 1 e 7  & 89,29 & 90,16 & 89,72 & \best{88,71} & 81,36 & 10\\ \hline \hline
      \textbf{Todas + NDVI\cite{torres_2021}} & {87,92}	&{90,46}	&{89,17}	&{87,99}	&{80,45}	&{61}	\\\hline 
      \textbf{RGB}   & 87,41	& 91,05	& 89,19	& 87,94	& 80,49 & 39   \\\hline
      \textbf{PCA} & \best{91,01} & 87,00 & 88,96 & 88,20 & 80,12 & 5  \\\hline

    \end{tabular}
    }
    \label{tab:semantica}
\end{table}

\section{Conclusões}
\label{sec:conclusao}
Neste trabalho, um novo arcabouço tem sido proposto para detecção de áreas de desmatamento da Amazônia. Por meio da seleção de bandas do satélite Landsat-8 baseada no algoritmo estocástico UMDA, foi possível encontrar as bandas que mais contribuem para a representação das áreas de interesse. Uma vez com esse conhecimento, as imagens da base foram reconstruídas utilizando apenas as bandas espectrais mais relevantes e essas imagens alimentaram uma arquitetura de segmentação semântica (DeepLabv3$+$) para tarefa de detecção de desmatamento. Nos experimentos realizados, foi possível mostrar que o arcabouço de seleção de bandas espectrais proposto pode ser uma solução para conseguir maior eficiência (redução de tempo de treinamento) e eficácia (qualidade de resultados) quando comparado com abordagens comumente utilizadas na literatura para a tarefa alvo.

\section*{Agradecimentos}
Os autores gostariam de agradecer a equipe do Projeto~\textit{ForestEyes}~\cite{foresteyes2019} pela colaboração, discussão e criação da base de dados de desmatamento. Além disso, agradecer a agência de fomento CNPq pela bolsa PIBIC do aluno Eduardo  B. Neto. Esta pesquisa é parte do INCT of the Future Internet for Smart Cities financiada pelo CNPq (processo \#465446/2014-0), Coordenação de Aperfeiçoamento de Pessoal de Nível Superior - Brasil (CAPES) - Código de Financiamento 001 e FAPESP (processos \#2018/23908-1, \#2019/26702-8, \#2014/50937-1 e \#2015/24485-9). Esse trabalho usou recursos do ``SDumont - Sistema de Computação Petaflópica do SINAPAD".



\begin{thebibliography}{10}
\providecommand{\url}[1]{#1}
\csname url@samestyle\endcsname
\providecommand{\newblock}{\relax}
\providecommand{\bibinfo}[2]{#2}
\providecommand{\BIBentrySTDinterwordspacing}{\spaceskip=0pt\relax}
\providecommand{\BIBentryALTinterwordstretchfactor}{4}
\providecommand{\BIBentryALTinterwordspacing}{\spaceskip=\fontdimen2\font plus
\BIBentryALTinterwordstretchfactor\fontdimen3\font minus \fontdimen4\font\relax}
\providecommand{\BIBforeignlanguage}[2]{{%
\expandafter\ifx\csname l@#1\endcsname\relax
\typeout{** WARNING: IEEEtran.bst: No hyphenation pattern has been}%
\typeout{** loaded for the language `#1'. Using the pattern for}%
\typeout{** the default language instead.}%
\else
\language=\csname l@#1\endcsname
\fi
#2}}
\providecommand{\BIBdecl}{\relax}
\BIBdecl

\bibitem{martin2015edge}
C.~Martin, \emph{On the {E}dge: {T}he {S}tate and {F}ate of the {W}orld's {T}ropical {R}ainforests}.\hskip 1em plus 0.5em minus 0.4em\relax Greystone Books Ltd, 2015.

\bibitem{nasaflorestas}
G.~Urquhart, W.~Chomentowski, D.~Skole, and C.~Barber, ``Tropical deforestation,'' 1998.

\bibitem{Hansen850}
{M. Hansen et al.}, ``High-resolution global maps of 21st-century forest cover change,'' \emph{science}, vol. 342, no. 6160, pp. 850--853, 2013.

\bibitem{prodes2022}
INPE, ``Estimativa de desmatamento na {A}mazônia {L}egal para $2022$ é de $11.568$ $km^2$,'' https://encurtador.com.br/aBDP2, 2022, accessed: 2023-08-01.

\bibitem{ortega2019evaluation}
M.~Ortega, J.~Bermudez, P.~Happ, A.~Gomes, and R.~Feitosa, ``Evaluation of deep learning techniques for deforestation detection in the amazon forest,'' \emph{ISPRS Annals of the Photogrammetry, Remote Sensing and Spatial Information Sciences}, vol.~4, pp. 121--128, 2019.

\bibitem{andrade2020evaluation}
{R. B. Andrade et al.}, ``Evaluation of semantic segmentation methods for deforestation detection in the amazon,'' \emph{ISPRS Archives; 43, B3}, vol.~43, no.~B3, pp. 1497--1505, 2020.

\bibitem{maretto2020spatio}
R.~V. Maretto, L.~M. Fonseca, N.~Jacobs, T.~S. K{\"o}rting, H.~N. Bendini, and L.~L. Parente, ``Spatio-temporal deep learning approach to map deforestation in amazon rainforest,'' \emph{IEEE Geoscience and Remote Sensing Letters}, vol.~18, no.~5, pp. 771--775, 2020.

\bibitem{yu2019selection}
Z.~Yu, L.~Di, R.~Yang, J.~Tang, L.~Lin, C.~Zhang, M.~S. Rahman, H.~Zhao, J.~Gaigalas, E.~G. Yu \emph{et~al.}, ``Selection of landsat 8 oli band combinations for land use and land cover classification,'' in \emph{2019 8th International Conference on Agro-Geoinformatics (Agro-Geoinformatics)}.\hskip 1em plus 0.5em minus 0.4em\relax IEEE, 2019, pp. 1--5.

\bibitem{foresteyes2019}
F.~Dallaqua, A.~Fazenda, and F.~Faria, ``{ForestEyes} {P}roject: {C}an {C}itizen {S}cientists {H}elp {R}ainforests?'' in \emph{{IEEE} 15th {International} {Conference} on {eScience}}.\hskip 1em plus 0.5em minus 0.4em\relax {IEEE}, 9 2019, pp. 18--27.

\bibitem{ma2003bandselection}
J.-P. Ma, Z.-B. Zheng, Q.-X. Tong, and L.-F. Zheng, ``An application of genetic algorithms on band selection for hyperspectral image classification,'' in \emph{Proceedings of the 2003 international conference on machine learning and cybernetics (IEEE Cat. No. 03EX693)}, vol.~5.\hskip 1em plus 0.5em minus 0.4em\relax IEEE, 2003, pp. 2810--2813.

\bibitem{nagasubramanian2018hyperspectral_ga}
K.~Nagasubramanian, S.~Jones, S.~Sarkar, A.~K. Singh, A.~Singh, and B.~Ganapathysubramanian, ``Hyperspectral band selection using genetic algorithm and support vector machines for early identification of charcoal rot disease in soybean stems,'' \emph{Plant methods}, vol.~14, pp. 1--13, 2018.

\bibitem{PCA}
I.~Jolliffe, \emph{Principal component analysis}.\hskip 1em plus 0.5em minus 0.4em\relax Springer, 2011.

\bibitem{SLIC}
{ R. Achanta et al.}, ``S{LIC} superpixels compared to state-of-the-art superpixel methods,'' \emph{IEEE transactions on pattern analysis and machine intelligence}, vol.~34, no.~11, pp. 2274--2282, 2012.

\bibitem{HARALICK}
{R. M. Haralick et al.}, ``Textural features for image classification,'' \emph{IEEE Transactions on systems, man, and cybernetics}, no.~6, pp. 610--621, 1973.

\bibitem{svm}
{B. E. Boser et al.}, ``A training algorithm for optimal margin classifiers,'' in \emph{Workshop on Computational Learning Theory}, ser. COLT '92, 1992, pp. 144--152.

\bibitem{muhlenbein1996recombination}
H.~M{\"u}hlenbein and G.~Paass, ``From recombination of genes to the estimation of distributions i. binary parameters,'' in \emph{International conference on parallel problem solving from nature}.\hskip 1em plus 0.5em minus 0.4em\relax Springer, 1996, pp. 178--187.

\bibitem{chen2018encoderdecoder}
{Liang-Chieh Chen et al.}, ``Encoder-decoder with atrous separable convolution for semantic image segmentation,'' 2018.

\bibitem{Ferreira_icpr2020}
{\'{A}}.~R. Ferreira, G.~H. de~Rosa, J.~P. Papa, G.~Carneiro, and F.~A. Faria, ``Creating classifier ensembles through meta-heuristic algorithms for aerial scene classification,'' in \emph{2020 25th International Conference on Pattern Recognition (ICPR)}, 2021, pp. 415--422.

\bibitem{deng2009imagenet}
{J. Deng et al.}, ``Imagenet: A large-scale hierarchical image database,'' in \emph{2009 IEEE conference on computer vision and pattern recognition}.\hskip 1em plus 0.5em minus 0.4em\relax Ieee, 2009, pp. 248--255.

\bibitem{torres_2021}
{D. Torres et al.}, ``Deforestation detection with fully convolutional networks in the amazon forest from landsat-8 and sentinel-2 images,'' \emph{Remote Sensing}, vol.~13, no.~24, 2021.

\end{thebibliography}
\end{document}